\documentclass[letterpaper, 10 pt, conference]{ieeeconf}
\IEEEoverridecommandlockouts
\usepackage{amsmath,amssymb,amsfonts}
\usepackage{array}
\usepackage{graphicx}
\usepackage{cite}
\usepackage{booktabs}
\usepackage{multirow}
\usepackage{makecell}
\usepackage{textcomp}
\usepackage{url}
\usepackage{stfloats}
\usepackage{xcolor}
\usepackage{tabularx}
\usepackage[hidelinks]{hyperref}

\newcolumntype{C}{>{\centering\arraybackslash}X}

\usepackage[caption=false,labelformat=simple]{subfig}

\newcommand{\btn}[1]{\texttt{#1}}
\newcommand{\st}[1]{\textsc{#1}}

\newcounter{remark}
\newenvironment{remark}[1]{%
  \par\smallskip\noindent
  \refstepcounter{remark}%
  \textbf{Remark~\theremark}\ \textit{(#1).}\enspace
}{%
  \par\smallskip
}

\newif\ifshowrevisions
\ifshowrevisions
  
\else
  
\fi

\begin{document}

\title{\LARGE \bf
MATE-BT: A Behavior-Tree-Based Framework for Hierarchically Decentralized Mission Autonomy in Heterogeneous Multi-Robot Teams}

\author{Sunbeen Lee$^{1}$, Hoseon Gi$^{1}$, Jik Hur$^{2}$, Inmo Jang$^{1*}$
\thanks{
This research was partially based on the results of a subcontracted project from Korea Aerospace Industries, LTD., conducted as part of the core technology development program led by the Korea Research Institute for Defense Technology Planning and Advancement.}
\thanks{
*Corresponding author: Inmo Jang (inmo.jang@kau.ac.kr)}
\thanks{$^{1}$Department of Artificial Intelligence, Korea Aerospace University, Goyang, Republic of Korea}
\thanks{$^{2}$Korea Aerospace Industries, Ltd., Sacheon, Republic of Korea}
}

\maketitle

\pagestyle{empty}
\thispagestyle{empty}

\begin{abstract}
Multi-robot teams executing predefined missions must adapt to runtime changes and sustain mission management despite robot failures.
This paper proposes MATE-BT (Mission, Allocation, and Task Execution with Behavior Trees), a hierarchically decentralized framework that separates mission execution, allocation, and task execution.
An active onboard leader manages mission progression and team-level task assignment, with authority transferable to another leader-capable robot after failure.
Interchangeable decentralized methods resolve target assignments, while each robot adapts its local behavior through planning and acting with behavior trees.
Webots simulations and real-robot experiments demonstrate mission continuity despite new targets and robot failures, including active-leader failure.
In the physical experiments, all MATE-BT computation runs onboard.
Hardware-in-the-loop resource measurements further support onboard execution on Jetson Orin Nano and Raspberry Pi~5.
By updating only the affected layers without full-plan regeneration, MATE-BT maintained a response time of approximately 130~ms with 30 robots and 120 targets.
Comparisons of four target-assignment plugins reveal scenario-dependent trade-offs between mission completion time and total travel distance.
\end{abstract}

\section{Introduction}
\label{sec:intro}

Heterogeneous teams combine platforms with different capabilities and responsibilities for mission management and task execution~\cite{Athanasiadou2026}.
In crewed--uncrewed operations, for example, onboard automation can support mission management on a crewed platform while uncrewed robots autonomously execute delegated tasks~\cite{ThomasFriedrich2026}.
Such teams must sustain predefined missions despite environmental changes, platform failures, and disrupted communication with external mission management.
This calls for decentralized mission autonomy based on onboard computation and inter-robot communication~\cite{Rizk2019}.

Runtime disturbances can require responses at different functional levels~\cite{Ingrand2017,Yan2013}.
At the task-execution level, a capable UGV may suppress a newly blocking fire before continuing toward its assigned victim, changing its local action sequence while retaining the assignment.
At the allocation level, a growing fire may require additional robots, or a robot failure may require unfinished work to be reassigned.
At the mission-execution level, the team must retain the ability to manage mission progression despite robot failures.
These cases motivate separating mission execution, allocation, and task execution so that disturbances can be addressed at the affected levels.

Prior work has addressed different aspects of these functional requirements.
For example, planning and acting with behavior trees (BTs), denoted \emph{PA-BT} in this paper, supports reactive robot-local behavior adaptation~\cite{Colledanchise2019}, while decentralized multi-robot task allocation (MRTA) methods coordinate assignments through local computation and communication~\cite{Chopra2017,Choi2009,Jang2018}.
Hierarchical multi-robot BTs provide a basis for organizing mission progression, allocation, and local execution~\cite{Colledanchise2016}.

We propose \textbf{MATE-BT (Mission, Allocation, and Task Execution with Behavior Trees)}, which connects these mechanisms in a layered onboard execution framework.
MATE-BT adopts a hierarchical form of decentralization: an active onboard leader manages mission progression and team-level task assignment, while target assignment and task execution are decentralized across robots; if the active leader fails, another leader-capable robot assumes mission-management authority.

\subsection{Related Work}
\label{sec:related}

This section reviews mission execution architectures for multi-robot autonomy, focusing on how mission management, task allocation, and robot-level execution are organized.
The architectures discussed below use BTs for execution, benefiting from reactive behavior through condition reevaluation and fallbacks, as well as modular composition and subtree reuse~\cite{Iovino2022,Colledanchise2017}.
The discussion here is organized by the responsibilities assigned to robots: (1) executing dispatched actions only; (2) managing local task execution under centralized coordination; and (3) participating in decentralized coordination decisions.

\subsubsection{Action execution from symbolic plans}
In monolithic architectures, mission management, task allocation, and task execution are encoded in a centrally generated symbolic plan, leaving each robot to execute only the actions dispatched to it.
\emph{PlanSys2}~\cite{Martin2021}, a ROS~2 successor inspired by \emph{ROSPlan}~\cite{Cashmore2015}, compiles a \emph{Planning Domain Definition Language} (PDDL) plan into a BT and dispatches its actions to robots.
When a change invalidates the plan, execution requires a plan-level update and regeneration.
The work in \cite{Rochefoucauld2025} extends this structure by partitioning execution among team-specific BT executors and selecting local replanning, robot reallocation, or full replanning according to failure impact.
Despite their different recovery scopes, both approaches rely on centrally generated symbolic plans that prescribe robot actions.

\subsubsection{Robot-local task execution}
These architectures assign complete tasks, rather than individual actions, and delegate their execution to robot-local BTs.
The HTN-based methods in~\cite{Neufeld2018,Rodrigues2022} derive agent-specific task plans through centralized planning or plan decomposition.
The framework in~\cite{Calvo2025} centrally optimizes assignments and schedules, and that in~\cite{Dahlquist2024} has a central unit select assignments from bids the robots estimate locally.
Despite these differences, authority over mission progression and allocation remains central, and each robot adapts only within the task it has been assigned.

In single robotic systems, \cite{DiPierno2025,Hallen2024} manage mission phases or operating modes using finite-state machines (FSMs) above local BT execution, but do not address decentralized coordination among independent robots.

\subsubsection{Distributed coordination among robots}
These architectures move at least part of the coordination authority to the robots, allowing them to participate in task allocation or delegation through local decision-making and inter-robot communication.
An early BT architecture in~\cite{Colledanchise2016} replicates global-task, task-assignment, and local-execution trees across robots, allowing each robot to compute mission progression and team-wide assignments.
However, it does not specify how conflicting decisions are reconciled when local information differs, a broader challenge in decentralization~\cite{JangDecReview}, and evaluates fault tolerance analytically without simulated or physical robot experiments.

\cite{Tadewos2023} retains a central mission controller but allows robots to auction prerequisite actions and extend their local BTs online.
The work in~\cite{Heppner2023} distributes designated subtree execution through utility-based delegation, while~\cite{Heppner2024} extends this idea with decentralized auctions for capability implementations and executing robots.
These approaches distribute task allocation or delegation, but either retain centralized mission management or do not specify how mission management continues after the loss of its executing component.

The studies reviewed in this subsection evaluate their proposed coordination mechanisms through analysis or simulation without reporting physical multi-robot experiments.
To our knowledge, no existing physical demonstration integrates onboard mission management, decentralized allocation, and robot-local execution while evaluating mission continuity after the failure of either a task-executing robot or the active mission manager.

\subsection{Contributions}
\label{sec:contributions}

\emph{1) Layered execution with hierarchical decentralization:}
MATE-BT separates mission execution, task allocation, and task execution so that a disturbance can be handled at the layer whose decision it affects.
The mission-management and team-level task-assignment functions are replicated across onboard leader-capable robots, allowing authority to transfer after active-leader failure while retaining the mission policy.
Target assignments are resolved through decentralized MRTA using local computation and inter-robot communication, while each robot adapts its behavior locally through PA-BT.

\emph{2) Interchangeable decentralized target assignment:}
A common plugin interface allows several established decentralized MRTA methods~\cite{Chopra2017,Choi2009,Jang2018} to be interchanged without changing either the mission-management or task-execution structure.
Controlled comparisons within each scenario show that plugin choice affects mission performance, with no method consistently preferred across the tested scenarios.

\emph{3) System-level evaluation of responsiveness and mission continuity:}
A comparison with a monolithic baseline shows that, by updating only the affected assignments instead of regenerating the full mission plan, MATE-BT handles newly added targets and limits response-time growth as mission scale increases.
Webots simulations and four-robot physical experiments demonstrate mission continuity after either a task-executing robot or the active leader fails.
In the physical experiments, all MATE-BT computation runs onboard, while central infrastructure supplies only the communication layer and detected-target information.
Resource measurements on Jetson Orin Nano and Raspberry Pi~5, spanning real-robot and hardware-in-the-loop tests, quantify the CPU and memory overhead of onboard BT execution.

\begin{figure*}[t]
\centering
 \includegraphics[width=\textwidth]{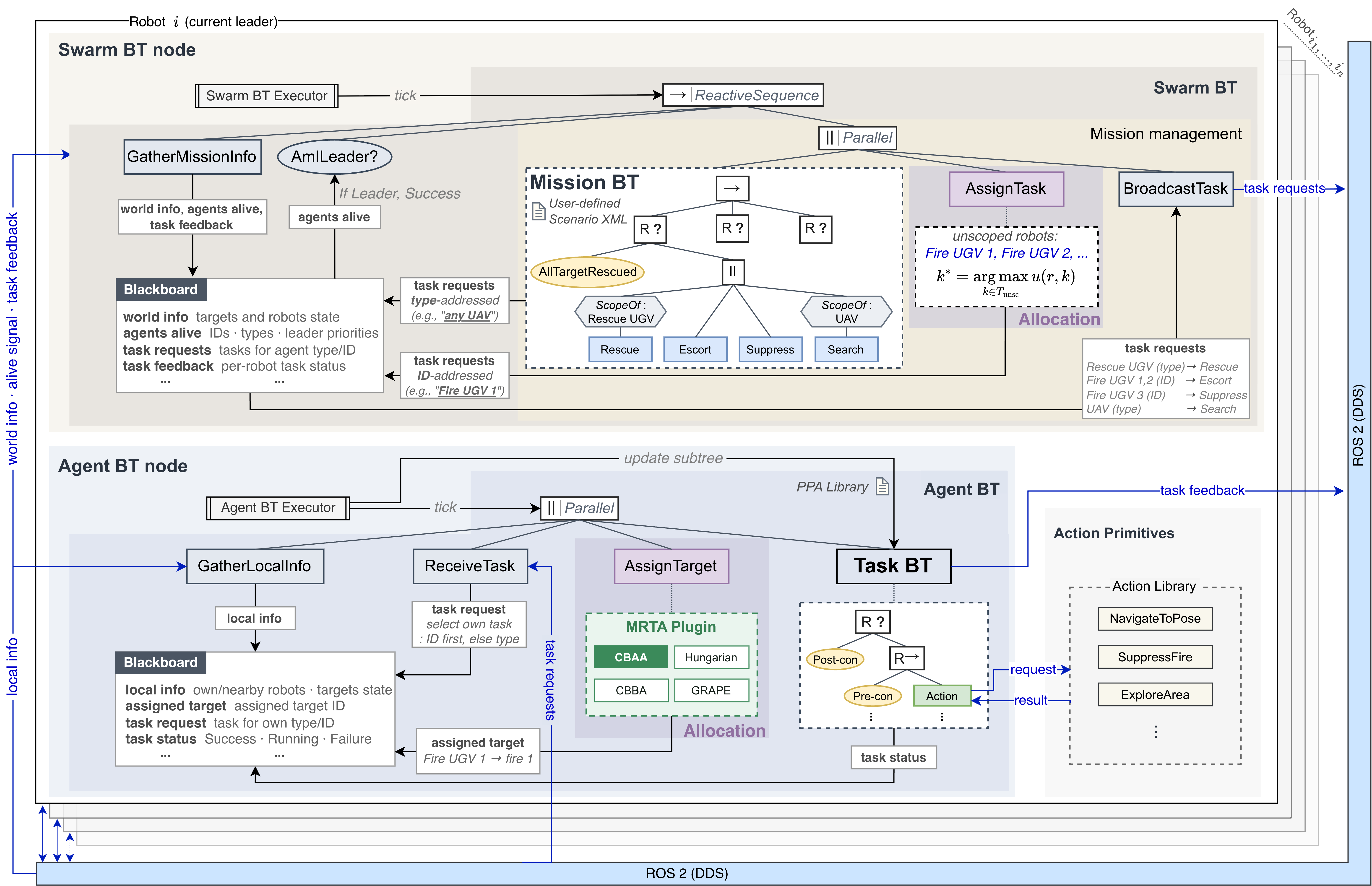}
\caption{Overall architecture of MATE-BT. The team-level Swarm BT runs on leader-capable robots, while the robot-level Agent BT runs on every robot. The Agent BT invokes robot-specific action primitives for task execution. Shaded blocks mark allocation, which spans both BTs: task assignment in the Swarm BT and target assignment in the Agent BT. Dashed blocks are supplied by the user for a given mission and robotic platforms. Blue lines are ROS interfaces between robots, and black arrows are blackboard reads and writes within a robot.}
\label{fig:architecture}
\end{figure*}

\section{Proposed Framework}
\label{sec:method}

\subsection{Problem Setting}
\label{sec:problem-setting}

MATE-BT assumes a \emph{mission policy} that is predefined yet abstract and structured as an ordered sequence of phases.
Each \emph{phase} specifies the tasks to be performed and the condition for phase completion.
The policy may also designate a preferred robot type for a task.
The mission policy is assumed to remain valid throughout execution.

As an example of such a policy, we use a fire-suppression-and-rescue mission throughout this paper.
First, in the \emph{Rescue phase}, UAVs search for victims, and Rescue UGVs transport detected victims to the base.
Fire UGVs escort the Rescue UGVs and clear fires obstructing the rescue routes.
They can also perform rescue when necessary but have a lower carrying capacity than Rescue UGVs.
This phase continues as new victims are detected and ends once all victims have been rescued.
Then, in the \emph{Suppress phase}, the Fire UGVs suppress the remaining fires until no active fires remain.
Finally, in the \emph{Return phase}, the surviving robots return to base, and the mission terminates once all have arrived.

We distinguish between a \emph{task} and a \emph{target} to describe the allocation decisions involved in a multi-robot mission.
A task denotes a type of work, such as search, rescue, escort, suppression, or return, whereas a target denotes the object or location to which that work applies, such as a victim, a fire, an escort position, or the base.
For a target-directed task, execution begins once both the task and its corresponding target have been assigned.

Each robot executes at most one task at a time, whereas a task request $k$ may be assigned to multiple robots and may specify an upper bound $n_k$ on their number.

During normal operation, we assume a connected, but not necessarily fully connected, inter-robot communication network.
Messages can be delivered over multiple hops when needed, although delays may cause temporary differences in robots' local information.

\subsection{Overall Architecture}

MATE-BT separates mission execution, allocation, and task execution into three functional layers.
Its implementation comprises two BT structures, a team-level \emph{Swarm BT} and a robot-level \emph{Agent BT}, together with robot-specific action primitives (Fig.~\ref{fig:architecture}).
The Swarm BT manages mission progression, while the Agent BT manages robot-local task execution through these primitives.
Allocation spans the two BTs: the Swarm BT assigns tasks to robots, while an MRTA plugin in the Agent BT assigns targets for those tasks.

Every robot runs an instance of the Agent BT onboard.
Leader-capable robots also run identical instances of the Swarm BT, allowing any of them to take over mission management; however, only the active leader ticks the mission-management subtree.

We implement BT executors that run the Swarm BT and Agent BT asynchronously in separate ROS~2 nodes.
Within each BT, nodes share information through a local blackboard.
Our ROS-aware leaf-node interfaces connect the BTs to ROS: condition nodes evaluate robot and world states received through topics, whereas action nodes invoke action primitives through ROS actions, services, or topic publications.
Task requests, execution feedback, and MRTA messages are exchanged among robots over ROS topics.

Each robot periodically publishes an alive signal containing its ID, robot type, and leader priority.
Swarm BT and Agent BT instances independently use these signals to maintain a local view of team membership for mission management and target assignment, respectively.
A robot is excluded from this view if no signal is received for a configured timeout $\tau$.

\subsection{Mission Execution}

The Swarm BT uses a \btn{ReactiveSequence} containing \btn{GatherMissionInfo}, \btn{AmILeader}, and a \btn{Parallel} subtree comprising the Mission BT, \btn{AssignTask}, and \btn{BroadcastTask} (Fig.~\ref{fig:architecture}).
\btn{GatherMissionInfo} first updates the local blackboard with world information, team information, and task-execution feedback.

\subsubsection{Mission Policy Encoding and Progression}

The \emph{Mission BT} encodes the mission policy as a \btn{Sequence} of phases, each represented by a \btn{ReactiveFallback} of its completion condition and a \btn{Parallel} of leaf nodes that issue task requests.
Each of these nodes records a required task and its parameters on the blackboard rather than invoking robot actions.
Each leaf returns \st{Success} once all robots assigned to its task have reported \st{Success}, and \st{Running} otherwise.
The phase-completion condition is reevaluated from world information whenever the phase is ticked.
A phase returns \st{Success} if and only if its completion condition holds or the configured number of task-request nodes have returned \st{Success}.
The \btn{Sequence} then advances to the next phase.

\subsubsection{Mission-Management Authority and Handover}

Leader-capable robots are ranked by configured priority, with ties broken by robot ID.
\btn{AmILeader} returns \st{Success} if and only if its robot is the highest-ranked candidate in its local membership view.
Robots with identical views therefore select the same leader.
When the active leader times out, the next-ranked candidate assumes mission-management authority.

\begin{remark}{Leader Convergence and Phase Recovery}
A communication partition may split the team into groups with different leaders.
After reconnection, the robots select a common leader once alive-signal exchange brings their membership views into agreement.
When a robot becomes the leader or its membership view expands, it resets its Mission BT \btn{Sequence} and reevaluates it from the first phase using current world information.
All phases in the evaluated mission define completion using such observable information.
\end{remark}

\subsection{Allocation}

\subsubsection{Task Assignment}

Task assignment determines which current-phase task each robot performs, while leaving target selection to the subsequent target-assignment stage.
MATE-BT resolves task assignment in two ways: \btn{ScopeOf} applies robot-type designations encoded in the mission policy, whereas \btn{AssignTask} assigns the remaining robots to unscoped tasks at runtime.

A \btn{ScopeOf} decorator addresses a task request exclusively to its designated robot type while that type remains available, and robots of that type are assigned only to the scoped task within the phase.
If no robot of the designated type remains, the task request becomes unscoped and is handled by \btn{AssignTask}.

At each tick, after scoped requests are resolved, \btn{AssignTask} assigns each robot $r$ in the remaining robot set $R_{\mathrm{unsc}}$ at most one unscoped task, while each task $k$ accepts at most $n_k$ robots.
The current implementation greedily processes pairs with $c_{r,k}>0$ in descending order of
\begin{equation}\label{eq}
u(r,k) = w_k \frac{c_{r,k}}{\max_{r' \in R_{\mathrm{unsc}}} c_{r',k}},
\end{equation}
where $w_k$ denotes task importance and $c_{r,k}$ the capability of robot $r$ for task $k$.
Accepted assignments are recorded as robot-specific task requests.

\btn{BroadcastTask} publishes the current task requests at every tick, and each robot's \btn{ReceiveTask} stores its assigned task and parameters in the local blackboard.

\subsubsection{Target Assignment}

Target assignment selects a target, such as a victim, fire, or destination, for each robot's assigned task.
To support interchangeable MRTA methods for this step, we define a standardized plugin interface following the common two-phase structure of decentralized MRTA: local decision-making and inter-robot conflict resolution~\cite{JangDecReview}.
At every tick, \btn{AssignTarget} in each Agent BT passes the available targets, the robot's state, and received messages to the configured plugin through this interface, then stores the selected target in the local blackboard.
The current implementation provides plugins for \emph{CBAA}, a greedy consensus-based auction, and its bundle extension \emph{CBBA}~\cite{Choi2009}, \emph{distributed Hungarian}~\cite{Chopra2017}, which converges to an optimal one-to-one assignment through local communication, and \emph{GRAPE}~\cite{Jang2018}, which forms coalitions through an anonymous hedonic game.
Collectively, these plugins cover the dominant MRTA problem classes involving instantaneous or time-extended assignment of single-robot-required tasks and instantaneous assignment of multi-robot-required tasks~\cite{Chakraa2023}.

\subsection{Task Execution}

\emph{Task BT} serves as a placeholder subtree in the Agent BT.
Using the task request and assigned target stored in the local blackboard, the Agent BT executor maps the current assignment to a desired postcondition and installs it as the child of Task BT whenever either changes.
The \emph{Postcondition--Precondition--Action (PPA) library} associates postconditions with the actions that achieve them and their required preconditions.
If the desired postcondition is unsatisfied, the executor applies PA-BT backchaining~\cite{Colledanchise2019} over this library to recursively add the required actions for the postcondition and any unmet preconditions.
Task BT executes the resulting subtree while its condition nodes reevaluate the latest blackboard state.
Task completion is determined by the observed postcondition rather than by an action result or service response, and the resulting status is reported as task feedback.

\begin{remark}{Application-Specific Components}
To instantiate MATE-BT, the application designer provides the components shown with dashed borders in Fig.~\ref{fig:architecture}: (i) a Mission BT encoding the mission policy, (ii) utility or cost models for task and target assignment and, if needed, a new target-assignment method implementing the standardized plugin interface, (iii) a PPA library for expanding the Task BT, and (iv) robot-specific action primitives and their ROS interfaces.
\end{remark}

\begin{figure*}[!t]
\centering

\includegraphics[width=\textwidth]{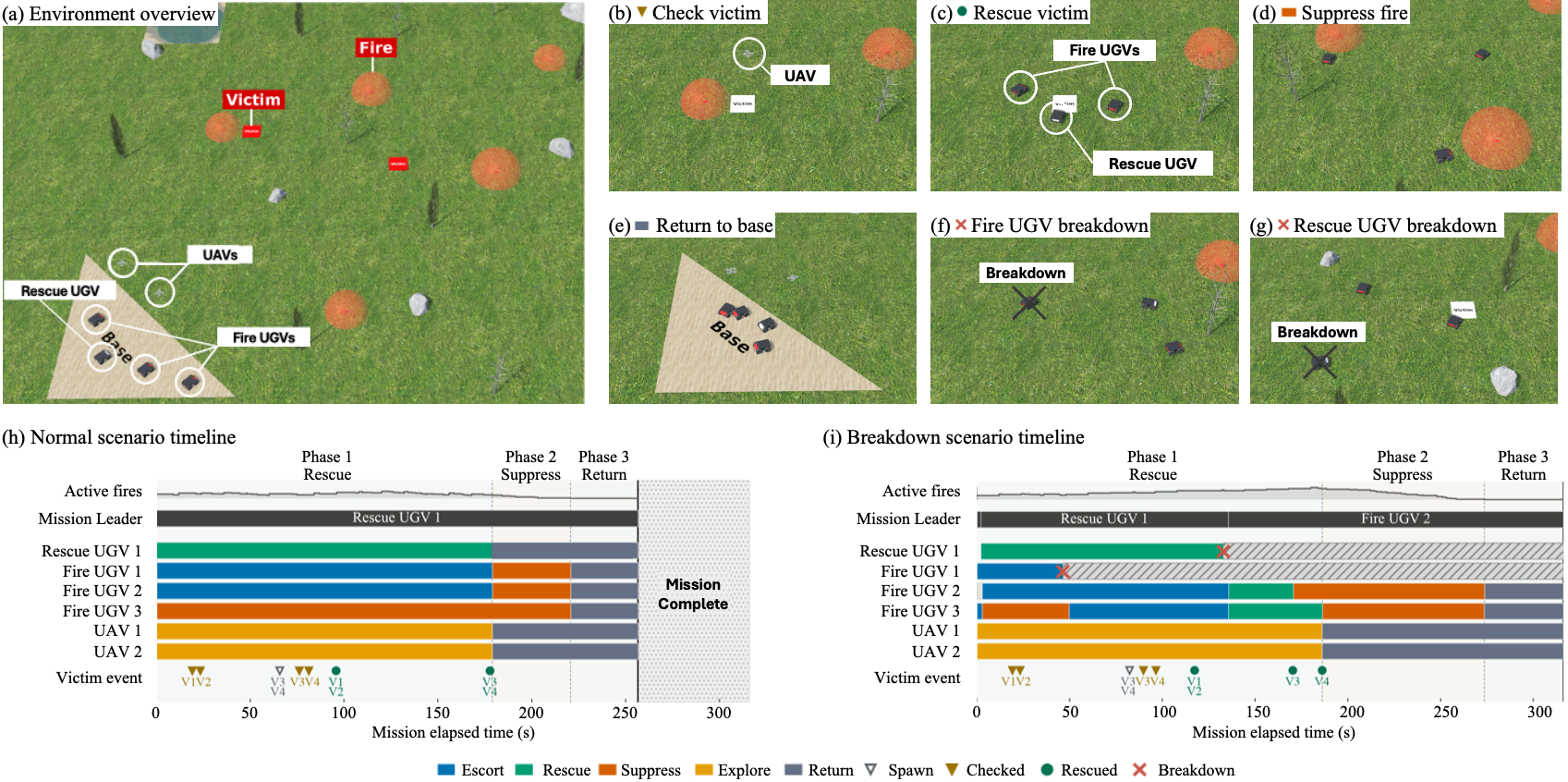}
\caption{Representative snapshots and mission timelines of the fire-suppression-and-rescue mission in Webots. Panel~(a) shows the environment and the robot team, panels~(b)–(e) show the normal mission sequence, and panels~(f) and~(g) show continued execution after a Fire UGV failure and a Rescue UGV failure, respectively. Panels~(h) and~(i) show the corresponding mission timelines for the normal and breakdown scenarios. Vertical dashed lines mark phase transitions, colored bars indicate robot tasks, the upper bands show the number of active fires and the mission-managing robot, and bottom markers denote victim events. In~(i), crosses mark robot breakdowns and hatched regions indicate subsequent robot unavailability.}
\label{fig:webots_snapshots}
\end{figure*}

\section{Experiments and Results}
\label{sec:experiments}

MATE-BT is implemented in ROS~2 Jazzy~\cite{Macenski2022} and evaluated through Webots simulations and real-robot experiments with AgileX LIMO robots\footnote{The video is available at \url{https://youtu.be/xHf-y_1Dl6M}.}.
The simulation and real-robot mission tests use the team configuration in Table~\ref{tab:mission-team}, with UAVs included only in simulation.
The physical team consists of four identical LIMO platforms configured as one Rescue UGV and three Fire UGVs.
All UGVs are leader-capable.
The Rescue UGV carries up to two victims at a time, while each Fire UGV carries one.

In the normal scenario, the team executes the mission policy described in Sec.~\ref{sec:problem-setting} as new victims and fires appear.
The breakdown scenario additionally introduces two robot failures during the Rescue phase.
Fire UGV~1 fails first while escorting the Rescue UGV.
The Rescue UGV, the active leader, then fails after delivering its victims to base.

The BTs tick at 10~Hz, alive signals are published at 5~Hz, and the alive-signal timeout $\tau$ is set to 2.5~s.
The simulation and real-robot mission tests use CBAA for target assignment.

\begin{table}
\caption{Heterogeneous Robot Team Configuration}
\label{tab:mission-team}
\centering
\footnotesize
\setlength{\tabcolsep}{3pt}
\begin{tabular*}{\columnwidth}{@{\extracolsep{\fill}}lccl@{}}
\toprule
Robot type & Quantity & Priority & Executable tasks \\
\midrule
Rescue UGV & 1 & 0    & rescue, return                   \\
Fire UGV   & 3 & 1--3 & suppress, escort, rescue, return \\
UAV        & 2 & --   & search, return          \\
\bottomrule
\end{tabular*}
\end{table}

\subsection{Simulation and Hardware-in-the-Loop Tests}

We evaluate mission continuity when new targets appear and either a task-executing robot or the active leader fails.
We also measure the CPU and memory overhead of onboard BT execution.

\begin{figure*}[!t]
\centering
\includegraphics[width=\textwidth]{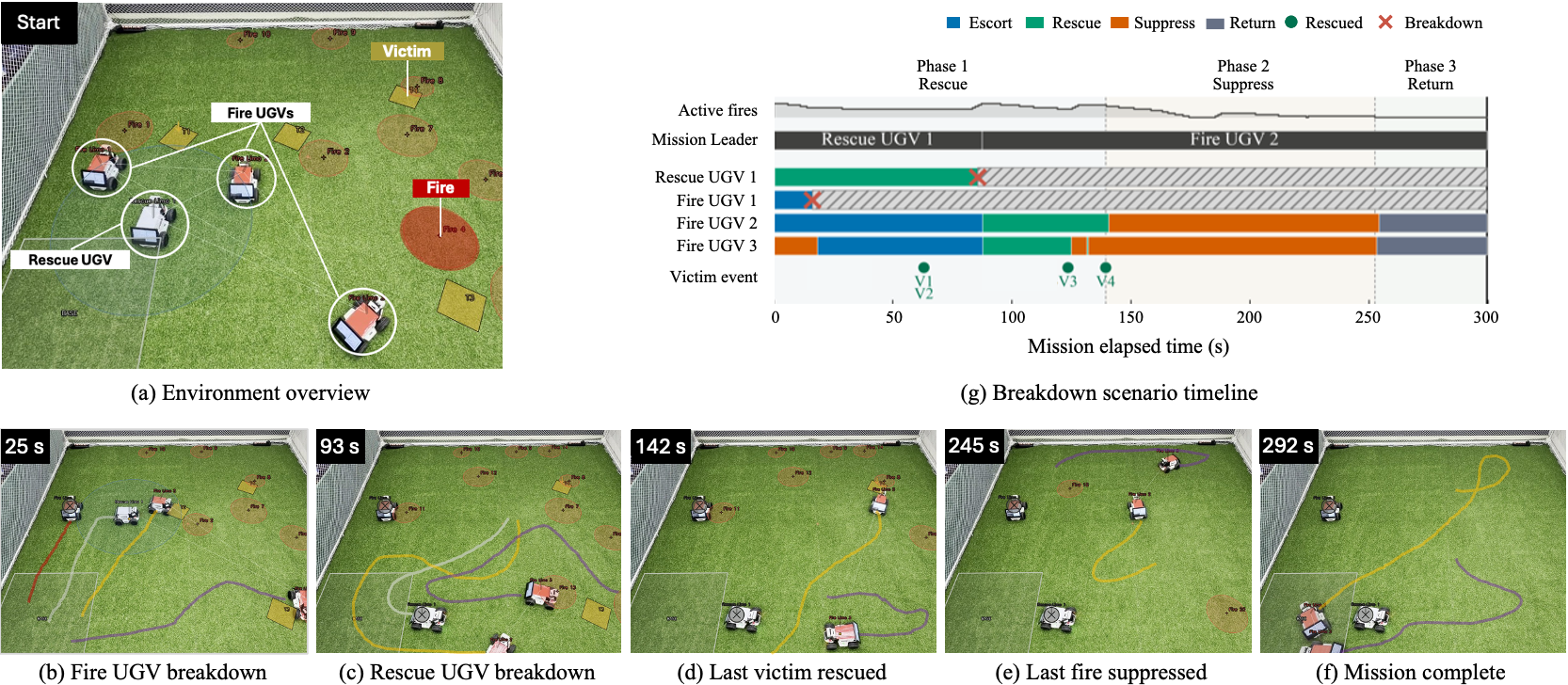}%
\caption{Breakdown scenario on four UGV robots, seen from the overhead
camera with augmented overlays. Red discs are fires, yellow squares are
victims, a lost robot carries a cross, and the base is at the lower left.
(a) The team starts the Rescue phase with Rescue UGV~1 escorted.
(b) Fire UGV~1 is lost.
(c) The leader, Rescue UGV~1, is lost.
(d) A Fire UGV rescues the last victim.
(e) The Suppress phase clears the remaining fires.
(f) The Return phase ends with the surviving robots at base.
(g) The mission timeline for this run, with the conventions of Fig.~\ref{fig:webots_snapshots}.}
\label{fig:limo_scenes}
\end{figure*}

In the normal scenario, the team completed the three phases in order while new victims and fires appeared during the mission~(Fig.~\ref{fig:webots_snapshots}(h)).
In the breakdown scenario~(Fig.~\ref{fig:webots_snapshots}(i)), \btn{AssignTask} reassigned the escort task to Fire UGV~3 after Fire UGV~1 failed.
When the subsequent failure of the active leader, Rescue UGV~1, was detected, Fire UGV~2, the next leader-capable robot in priority order, took over mission management and regenerated the current-phase task requests.
With no Rescue UGV remaining, \btn{AssignTask} assigned rescue to both surviving Fire UGVs according to their utilities, while robots with unaffected assignments continued their tasks.
The team completed both scenarios without operator intervention.

\begin{table}[t]
\caption{Comparison of Failure Responses Across Scenarios}
\label{tab:failure-response}
\centering
\footnotesize
\setlength{\tabcolsep}{3pt}
\begin{tabular*}{\columnwidth}{@{\extracolsep{\fill}}l
    *{3}{r@{\extracolsep{0pt}\,}l@{\extracolsep{\fill}}}}
\toprule
Metric & \multicolumn{2}{c}{Normal} & \multicolumn{2}{c}{Breakdown 1} & \multicolumn{2}{c}{Breakdown 2} \\
\midrule
Runs completed       & \multicolumn{2}{c}{10/10} & \multicolumn{2}{c}{10/10} & \multicolumn{2}{c}{10/10} \\
Completion time~(s)  & 252.74 & {\tiny$\pm$12.28} & 323.61 & {\tiny$\pm$42.32} & 366.92 & {\tiny$\pm$33.63} \\
Handover latency~(s) & \multicolumn{2}{c}{--} & 2.44 & {\tiny$\pm$0.08} & 2.46 & {\tiny$\pm$0.06} \\
Task reassignments   & 0.0 & {\tiny$\pm$0.0} & 4.0 & {\tiny$\pm$0.0} & 1.0 & {\tiny$\pm$0.0} \\
Target reassignments & 63.4 & {\tiny$\pm$6.5} & 77.6 & {\tiny$\pm$15.6} & 85.6 & {\tiny$\pm$10.5} \\
Task BT expansions   & 37.4 & {\tiny$\pm$0.5} & 51.2 & {\tiny$\pm$1.6} & 40.0 & {\tiny$\pm$0.7} \\
\bottomrule
\end{tabular*}
\vspace{3pt}
\parbox{\columnwidth}{\raggedright\scriptsize
   Mean $\pm$ s.d. over ten runs per scenario. Both breakdown scenarios include the failure of Fire UGV~1, and the active leader fails at a randomized time during the Rescue phase in Breakdown~1 and while at base during the Suppress phase in Breakdown~2.}
\end{table}

To quantify these responses across repeated runs, we compared the normal scenario with two breakdown scenarios over ten runs each using the same random seeds (Table~\ref{tab:failure-response}).
Both breakdown scenarios included Fire UGV~1's failure during the Rescue phase.
The active leader, Rescue UGV~1, subsequently failed at a randomized time during Rescue in Breakdown~1 or while at base during Suppress in Breakdown~2.

Mean handover latencies, measured from leader failure until the successor identified itself as leader, were 2.44 and 2.46~s, respectively.
These were slightly below $\tau=2.5$~s because the timeout started from the last received alive signal, before the failure.
Task reassignments count task changes within a phase, excluding phase transitions.
Breakdown~1 recorded four: one switch to escort after Fire UGV~1 failed, two to rescue after Rescue UGV~1 failed, and one back to suppression after a robot completed its rescue work.
Breakdown~2 recorded only the escort reassignment, as leader failure left the ongoing suppression assignments unchanged.
Task BT expansions increased more in Breakdown~1, consistent with the additional expansion needed to execute the new rescue assignments.
Both breakdown scenarios had longer mission durations and more target reassignments than normal.
Normalizing mean counts by mean completion times yielded similar rates of approximately 14--15 reassignments per minute across all scenarios, suggesting that the higher totals partly reflected longer execution.



We also measured onboard resource usage through hardware-in-the-loop simulation~(HILS).
In each run, a Raspberry Pi~5 or a Jetson Orin Nano executed the BTs of one robot, while an external PC ran Webots, the low-level controllers, and the BTs of the remaining robots.
Each board was evaluated as an active leader, a non-leading leader-capable robot, and an Agent-BT-only robot.

Across these roles, CPU utilization relative to the idle reference increased by 19.9 to 22.5 percentage points on the Raspberry Pi~5 and by 14.3 to 15.3 percentage points on the Jetson Orin Nano, while RAM utilization increased by 1.5 to 2.1 and 0.9 to 2.0 percentage points, respectively~(Table~\ref{tab:resource}).
These measurements support the feasibility of running both trees onboard the tested platforms.

\begin{table}[t]
\caption{Onboard Resource Utilization per Role}
\label{tab:resource}
\centering
\footnotesize
\setlength{\tabcolsep}{3.5pt}
\begin{tabular*}{\columnwidth}{@{\extracolsep{\fill}}lrrrr@{}}
\toprule
                 & \multicolumn{2}{c}{Raspberry Pi 5}
                 & \multicolumn{2}{c}{Jetson Orin Nano} \\
\cmidrule(lr){2-3}
\cmidrule(lr){4-5}
Role             & CPU (\%) & RAM (MiB) & CPU (\%) & RAM (MiB) \\
\midrule
\textit{Idle (ref.)}
    & \textit{6.8} & \textit{1419} & \textit{5.4} & \textit{1851} \\
\midrule
\multirow{2}{*}{\makecell[l]{Active leader\\\scriptsize (Swarm BT + Agent BT)}}
    & 29.3 & 1586 & 20.7 & 2006 \\
    & \scriptsize(+22.5) & \scriptsize(+2.1) & \scriptsize(+15.3) & \scriptsize(+2.0) \\
\addlinespace
\multirow{2}{*}{\makecell[l]{Leader-capable\\\scriptsize (Swarm BT + Agent BT)}}
    & 28.4 & 1534 & 20.2 & 1952 \\
    & \scriptsize(+21.6) & \scriptsize(+1.5) & \scriptsize(+14.8) & \scriptsize(+1.3) \\
\addlinespace
\multirow{2}{*}{\makecell[l]{Others\\\scriptsize (Agent BT only)}}
    & 26.7 & 1564 & 19.7 & 1916 \\
    & \scriptsize(+19.9) & \scriptsize(+1.8) & \scriptsize(+14.3) & \scriptsize(+0.9) \\
\bottomrule
\end{tabular*}
\vspace{3pt}
\parbox{\columnwidth}{\raggedright\scriptsize
The reported values are averages over ten HILS runs for each board and role.
Parentheses indicate percentage-point~(\%p) increases relative to the idle reference for each board.}
\end{table}

\subsection{Real-Robot Tests}

The real-robot tests use the same Mission BT as the simulation tests.
Each robot ran the Swarm BT, the Agent BT, and its driving stack on a mounted Jetson Orin Nano.
Central infrastructure provided only the Zenoh-based communication layer and detected-target information.

The team autonomously completed both scenarios indoors~(Fig.~\ref{fig:limo_scenes}).
The breakdown test reproduced the leader handover and task reassignment observed in simulation.

\begin{figure*}[!t]
\centering
\begin{minipage}[t]{0.32\textwidth}
\centering
\captionsetup[subfloat]{labelformat=empty}
\subfloat[\vphantom{Reactivity test}]{\includegraphics[width=\linewidth]{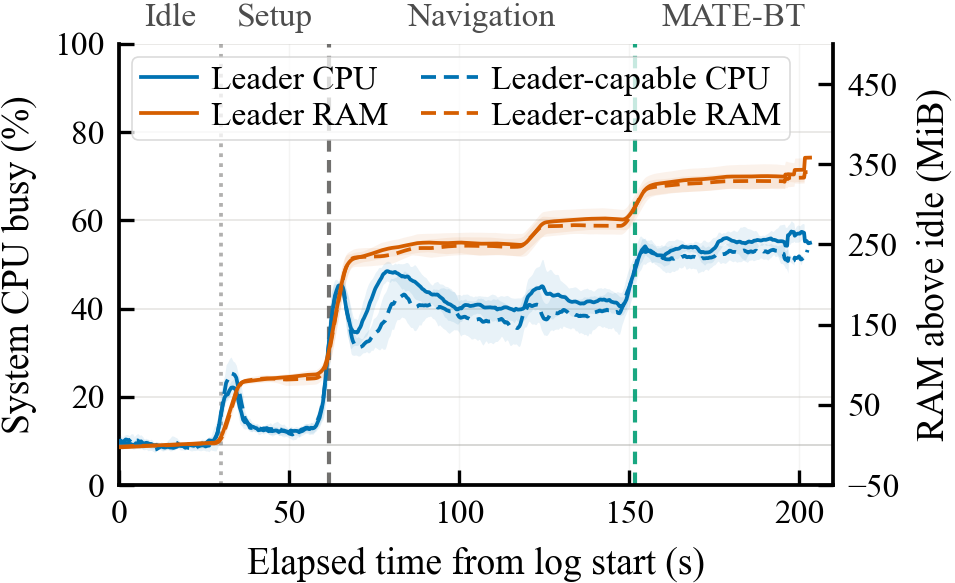}}
\caption{CPU (left axis) and RAM (right axis) utilization, averaged over ten runs. Non-leader denotes a leader-capable robot that is not leading. The grey and green lines mark the start of the driving stack and of MATE-BT.}
\label{fig:limo_resource}
\end{minipage}\hfill
\begin{minipage}[t]{0.65\textwidth}
\centering
\subfloat[Reactivity test]{%
 \includegraphics[width=0.48\linewidth]{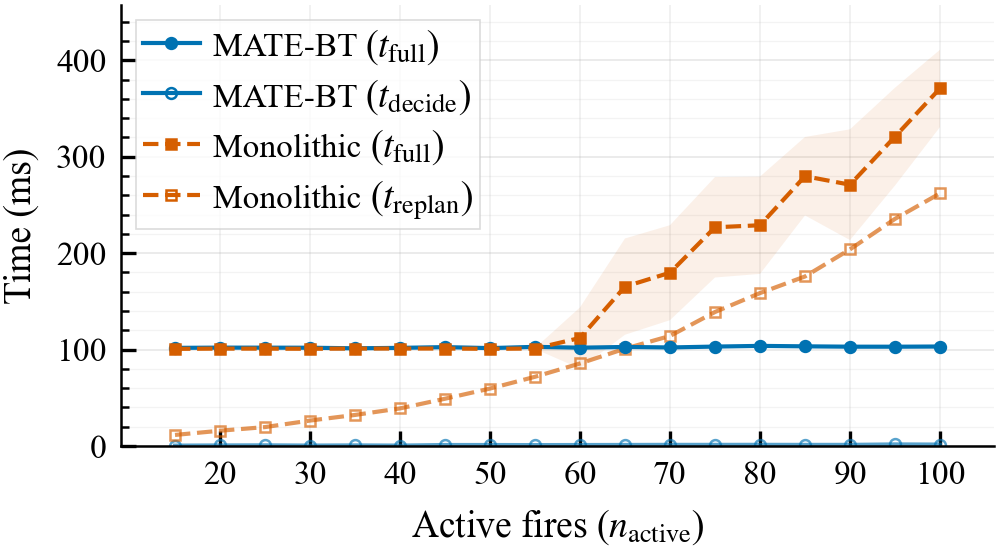}%
  \label{fig:reactivity}}
  \hfill
\subfloat[Scalability test]{%
  \includegraphics[width=0.48\linewidth]{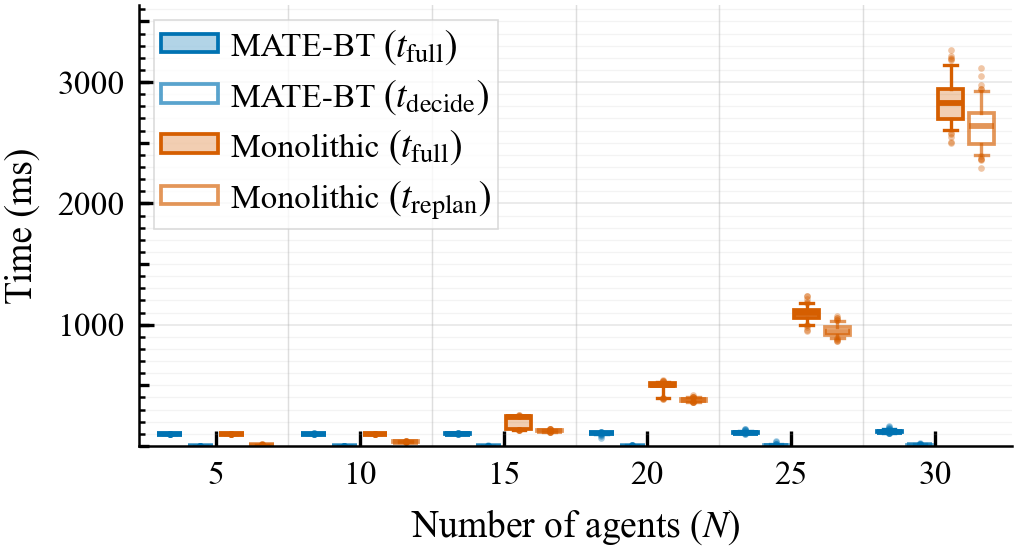}%
  \label{fig:scalability}}
\caption{Reactivity and scalability of MATE-BT and the monolithic baseline, measured over ten repetitions per condition.
(a) Mean $t_{\mathrm{full}}$, $t_{\mathrm{decide}}$, and $t_{\mathrm{replan}}$ as active fires increase, with bands spanning one standard deviation around mean $t_{\mathrm{full}}$.
(b) Distribution of the same metrics as the numbers of robots and fire targets increase, with boxes for the mean and interquartile range and whiskers at the 5th and 95th percentiles.}
\label{fig:reactivity_scalability}
\end{minipage}
\end{figure*}

Figure~\ref{fig:limo_resource} shows onboard resource usage during the idle, setup, navigation, and MATE-BT execution stages.
Communication and coordinate transforms were initialized during setup, followed by the Nav2-based driving stack~\cite{Macenski2020} for navigation and the Swarm BT and Agent BT for MATE-BT execution.
Enabling MATE-BT increased system CPU and RAM utilization by approximately 13 and 2 percentage points, respectively, with little difference between the active leader and a non-leading leader-capable robot.
Resource usage then remained stable throughout mission execution.

\subsection{Reactivity and Scalability Test}

We evaluate responsiveness to new targets and scalability with team size by comparing MATE-BT with a monolithic planning baseline.
Both tests use only the Suppress phase.
We implemented the baseline with a planning and execution pipeline inspired by PlanSys2~\cite{Martin2021}.
Given a PDDL model of the scenario, the baseline generates a plan using POPF~\cite{Coles2010} and converts it into a single BT that dispatches actions to the robots.
Both methods ran in the same Webots environment with common ROS state interfaces and a 100~ms tick period.
To match the baseline's centralized setup, MATE-BT's target assignment was run centrally with the Hungarian algorithm in this comparison only.
The robots remained stationary throughout both tests to focus on computational responsiveness.

Our primary comparison uses $t_{\mathrm{full}}$, measured from the detection of a new fire or a scheduled update trigger until an action goal reflecting the updated assignment is issued.
As supplementary measurements, we record target-assignment computation time in MATE-BT ($t_{\mathrm{decide}}$) and plan regeneration time in the baseline ($t_{\mathrm{replan}}$).
Task BT expansion before goal dispatch is included in $t_{\mathrm{full}}$ but excluded from $t_{\mathrm{decide}}$.

\subsubsection{Reactivity Test}

Ten Fire UGVs were deployed, and five new fires were added every 5~s, increasing the active fire count from 10 to 100.
As shown in Fig.~\ref{fig:reactivity_scalability}\subref*{fig:reactivity}, MATE-BT's $t_{\mathrm{full}}$ remains near 100~ms.
Its much shorter $t_{\mathrm{decide}}$ suggests that this response time is largely determined by the BT tick period.
For the baseline, $t_{\mathrm{full}}$ reaches approximately 370~ms at 100 fires, while $t_{\mathrm{replan}}$ increases with the fire count and exceeds the tick period around 60 fires.
The monolithic baseline updates and re-solves the PDDL problem when new fires appear, incurring greater replanning costs as the fire count grows.
MATE-BT instead updates the target set and assignments while retaining the Mission BT.

\subsubsection{Scalability Test}

Team size ranges from 5 to 30 robots, with four times as many fire targets.
Because the target set remains fixed for each team size, both methods receive an update trigger every 5~s.
MATE-BT's $t_{\mathrm{full}}$ remains near the tick period, reaching approximately 130~ms with 30 robots and 120 fire targets.
The baseline's response time was comparable up to 10 robots but increased to approximately 2,800~ms at 30 robots (Fig.~\ref{fig:reactivity_scalability}\subref*{fig:scalability}).

\subsection{Allocation Strategy Evaluation}

This experiment evaluates how MRTA plugin choice for target assignment affects mission performance.
CBAA, distributed Hungarian, CBBA, and GRAPE are compared without modifying the Mission BT or Agent BT.

\begin{figure}[t]
\centering
\includegraphics[width=\columnwidth]{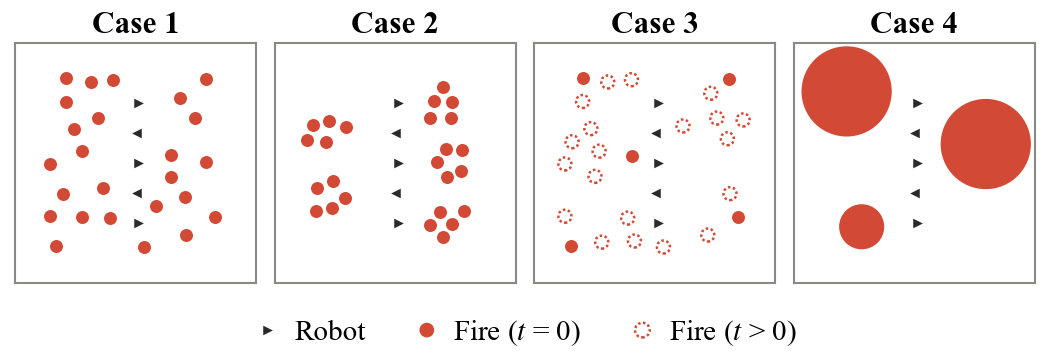}
\caption{Allocation cases with spread fires in Case~1, clustered fires in Case~2, sequential fire batches in Case~3, and large fires requiring cooperation in Case~4.}
\label{fig:allocation_cases}
\end{figure}

All four cases use the Suppress phase and the same team of five Fire UGVs (Fig.~\ref{fig:allocation_cases}).
Cases~1 and~2 each contain 25 fires in spread and clustered configurations, respectively.
Case~3 introduces five batches of five fires, with the next batch appearing immediately after the previous one is fully suppressed.
Case~4 contains three large fires to evaluate cooperative suppression.
Fire positions are randomized for each episode.
To model the benefit of cooperation, the suppression rate at each fire is proportional to the number of robots working on it.
Each plugin is evaluated over 15 episodes per case using mission completion time and total travel distance.

Table~\ref{tab:allocation_results} shows similar mean completion times across methods in Case~1.
CBBA, distributed Hungarian, and GRAPE achieve the lowest mean completion times in Cases~2, 3, and~4, respectively.
CBBA also yields the shortest mean total travel distances in Cases~1--3.
In these runs, its bundles allow some robots to handle successive fires while robots without assigned targets wait, concentrating movement on fewer robots.
GRAPE benefits from cooperative suppression in Case~4 but incurs the largest mean travel distance in every case as its coalitions move between fires.
No plugin performs best across all cases and both metrics.

\begin{table}
\caption{MRTA Plugin Comparison Across Cases}
\label{tab:allocation_results}
\centering
\scriptsize
\setlength{\tabcolsep}{3pt}
\renewcommand{\arraystretch}{1.0}
\begin{tabularx}{\linewidth}{@{}c l C C C C@{}}
\toprule
 & Allocator & Case 1 & Case 2 & Case 3 & Case 4 \\
\midrule
\multirow{4}{*}{\rotatebox[origin=c]{90}{Time (s)}}
 & CBAA      & 103.54\,{\tiny$\pm$11.54} & 90.36\,{\tiny$\pm$15.93} & 178.34\,{\tiny$\pm$18.45} & 86.99\,{\tiny$\pm$1.86} \\
 & Dist.\ Hungarian & 102.88\,{\tiny$\pm$7.48}  & 93.62\,{\tiny$\pm$16.50} & \textbf{155.84\,{\tiny$\pm$10.14}} & 85.91\,{\tiny$\pm$1.99} \\
 & CBBA      & \textbf{101.13\,{\tiny$\pm$8.53}} & \textbf{83.70\,{\tiny$\pm$7.99}} & 181.49\,{\tiny$\pm$12.57} & 86.96\,{\tiny$\pm$1.67} \\
 & GRAPE     & 103.92\,{\tiny$\pm$12.01} & 105.86\,{\tiny$\pm$13.89} & 227.24\,{\tiny$\pm$33.06} & \textbf{60.11\,{\tiny$\pm$4.89}} \\
\midrule
\multirow{4}{*}{\rotatebox[origin=c]{90}{Dist.\ (m)}}
 & CBAA      & 313.97\,{\tiny$\pm$36.22} & 265.18\,{\tiny$\pm$87.00} & 453.74\,{\tiny$\pm$35.66} & 46.12\,{\tiny$\pm$5.79} \\
 & Dist.\ Hungarian & 337.74\,{\tiny$\pm$49.21} & 292.12\,{\tiny$\pm$93.39} & 478.55\,{\tiny$\pm$48.58} & \textbf{44.96\,{\tiny$\pm$5.02}} \\
 & CBBA      & \textbf{243.23\,{\tiny$\pm$19.64}} & \textbf{157.05\,{\tiny$\pm$19.75}} & \textbf{350.55\,{\tiny$\pm$29.42}} & 46.06\,{\tiny$\pm$5.73} \\
 & GRAPE     & 396.58\,{\tiny$\pm$68.27} & 377.38\,{\tiny$\pm$76.80} & 1040.10\,{\tiny$\pm$191.20} & 112.83\,{\tiny$\pm$26.24} \\
\bottomrule
\end{tabularx}
\vspace{3pt}
\parbox{\columnwidth}{\raggedright\scriptsize
Mission-completion time and total travel distance, mean $\pm$ s.d. over 15 episodes per case. Best per case in bold.}
\end{table}

\section{Conclusion}

This paper presented MATE-BT, a behavior-tree-based hierarchically decentralized framework for mission autonomy in heterogeneous multi-robot teams.
It combines transferable onboard mission management, interchangeable decentralized MRTA plugins, and reactive robot-local execution while retaining a predefined mission policy.
Webots simulations and real-robot experiments demonstrated mission continuity despite new targets and robot failures, including active-leader failure.
All MATE-BT computation ran onboard in the real-robot experiments.
Hardware-in-the-loop resource measurements further supported onboard execution on Jetson Orin Nano and Raspberry Pi~5.
Compared with the monolithic baseline, which regenerates its full plan, MATE-BT maintained response times near the tick period with up to 30 robots and 120 targets.

Applying MATE-BT to a new mission currently requires manually specifying the Mission BT and the PPA library.
Future work could automate the generation and verification of these components using mission requirements and robot capabilities.
The observed variation in plugin performance motivates automatic MRTA plugin selection or a single method that adapts to different mission conditions.
Extending plugin-based configuration to \btn{AssignTask} would allow both task-assignment and target-assignment methods to be selected without modifying the framework.

\bibliographystyle{IEEEtran}
\bibliography{paper}
\end{document}